\pdfoutput=1
\PassOptionsToPackage{dvipsnames,table}{xcolor}
\documentclass[letterpaper]{article} 
\usepackage{aaai2027}  

\usepackage[hyphens]{url}  
\usepackage{graphicx} 
\usepackage{natbib}  
\usepackage{caption} 
\usepackage{algorithm}
\usepackage{algpseudocode}
\usepackage{amsmath}
\usepackage{tcolorbox}
\usepackage{amssymb}
\usepackage{booktabs}
\definecolor{mycolor}{RGB}{211,211,211} 
\usepackage{makecell}
\usepackage{newfloat}
\usepackage{listings}

\DeclareCaptionStyle{ruled}{labelfont=normalfont,labelsep=colon,strut=off} 
\floatstyle{ruled}
\newfloat{listing}{tb}{lst}{}
\floatname{listing}{Listing}

\title{Mind What Matters for Reasoning: Aligning Cross-Modal Attention via Selective Probability Mass Concentration}

\author{
    Jiaqi Deng\textsuperscript{\rm 1},
    Zonghan Wu\textsuperscript{\rm 2}, Zhan Heng\textsuperscript{\rm 3}, Xiaoshui Huang\textsuperscript{\rm 4}, Huan Huo\textsuperscript{\rm 1}\corresponding, Guandong Xu\textsuperscript{\rm 5}\corresponding
}
\affiliations{
    \textsuperscript{\rm 1}University of Technology Sydney
    \textsuperscript{\rm 2}East China Normal University
    \textsuperscript{\rm 3}The University of New South Wales\\
    \textsuperscript{\rm 4}Shanghai Jiaotong University
    \textsuperscript{\rm 5}
The Education University of Hong Kong
    
}

\begin{document}

\maketitle

\begin{abstract}
   Multimodal large language models (MLLMs) achieve strong performance on visual reasoning tasks, yet remain prone to hallucinations and over-reliance on language priors, often generating answers without adequately using task-relevant visual evidence. Existing approaches primarily improve reasoning through reasoning-oriented supervision or inference-time strategies. In this work, we study a complementary question: \textit{can multimodal reasoning be improved by strengthening implicit visual grounding} without directly supervising the reasoning process? Motivated by the functional specialization of attention heads, we investigate whether reasoning can be improved by guiding only the heads most responsive to visual evidence grounding. We propose Selective Probability Mass Concentration (sPMC), a training framework that identifies grounding-responsive heads and selectively regularizes their text-to-image attention. sPMC treats normalized attention over visual tokens as a spatial probability distribution and encourages the probability mass to be assigned to semantically relevant regions using segmentation-derived spatial priors. Adaptive Head Selection restricts this guidance to visually responsive heads while leaving the remaining heads unconstrained to preserve their complementary functions. Across 6 multimodal benchmark suites, sPMC achieves an average zero-shot improvement of 3\% and gains of up to 11.3\% across multiple MLLMs while regularizing only 3\%-15\% of their attention heads. These results demonstrate that targeted guidance of sparse and implicit visual evidence pathways can directly improve multimodal reasoning.

\end{abstract}



\maketitle

\section{Introduction}
\label{sec:intro}
Multimodal large language models (MLLMs)~\cite{Bai2023Qwen-VL:Beyond,Dai2023InstructBLIP:Tuning,Team2024Chameleon:Models} have made substantial progress on visually grounded reasoning tasks, including Visual Question Answering (VQA)~\cite{Li2025VoCoT:Models,Gao2025Interleaved-ModalChain-of-Thought,Deng2025ATask}. Nevertheless, they remain prone to hallucinations and over-reliance on language priors~\cite{Leng2024MitigatingDecoding,Bai2024HallucinationSurvey}, often producing answers without adequately using task-relevant visual evidence. This problem is especially acute in fine-grained reasoning, where a correct answer may depend on only a few informative visual cues.

Recent analyses associate these failures with misallocated text-to-image attention~\cite{Seil2025SeeModels,Woo2025DontModels,Chen2025Lavender:Tuning}. Attention determines which visual signals are emphasized and propagated during decoding; when it is diffuse or concentrated on irrelevant regions, the model may aggregate incomplete or misleading evidence. These findings suggest that reasoning failures are often linked not only to deficiencies in reasoning itself, but also to ineffective utilization of visual evidence. 

Existing methods therefore attempt to improve visual grounding either by manipulating attention dynamics during inference~\cite{Seil2025SeeModels,Woo2025DontModels,Leng2024MitigatingDecoding} or by directly aligning attention maps during training~\cite{Chen2025Lavender:Tuning}. While inference-time approaches can partially mitigate attention misallocation, they operate on the model's native attention dynamics and do not fundamentally improve how visual evidence is utilized. Training-based approaches provide stronger supervision signals, but rigidly assume dense alignment across attention maps, requiring broad modification of the attention mechanism. Despite their methodological differences, both paradigms implicitly treat attention as a uniformly important component. Recent work suggests that visual grounding is concentrated in a small subset of attention heads~\cite{Seil2025SeeModels,Kang_2025_CVPR}. This raises a unexplored question: \textit{can multimodal reasoning be improved by selectively strengthening the attention pathways exhibiting strong visual-grounding signals, rather than directly the reasoning process itself?}

In response, we propose \textbf{Selective Probability Mass Concentration (sPMC)}, a training framework for targeted text-to-image attention guidance. Given a segmentation-derived spatial mask, sPMC increases the total attention mass assigned to the masked region without prescribing its distribution across individual visual tokens. Because the objective constrains only aggregate in-mask mass, it tolerates moderately over-inclusive masks without requiring every masked token to match a fixed target. Through a head-level empirical study, we confirm that strong visual engagement and spatial grounding are consistently concentrated in a small subset of attention heads. Building on this finding, \textbf{Adaptive Head Selection} applies sPMC only through these grounding-responsive heads while leaving the remaining heads unconstrained. Controlled ablations further show that guiding this subset yields consistent improvements in downstream reasoning and is more effective than all-head guidance. This observation suggests that effective visual evidence utilization is not uniformly distributed throughout the attention mechanism. Instead, a relatively small number of attention heads appear to form attention pathways that exhibit particularly strong visual-grounding behavior. Our contributions are:
\begin{figure*}[ht]
    \centering \includegraphics[width=.9\textwidth, height=160 pt]{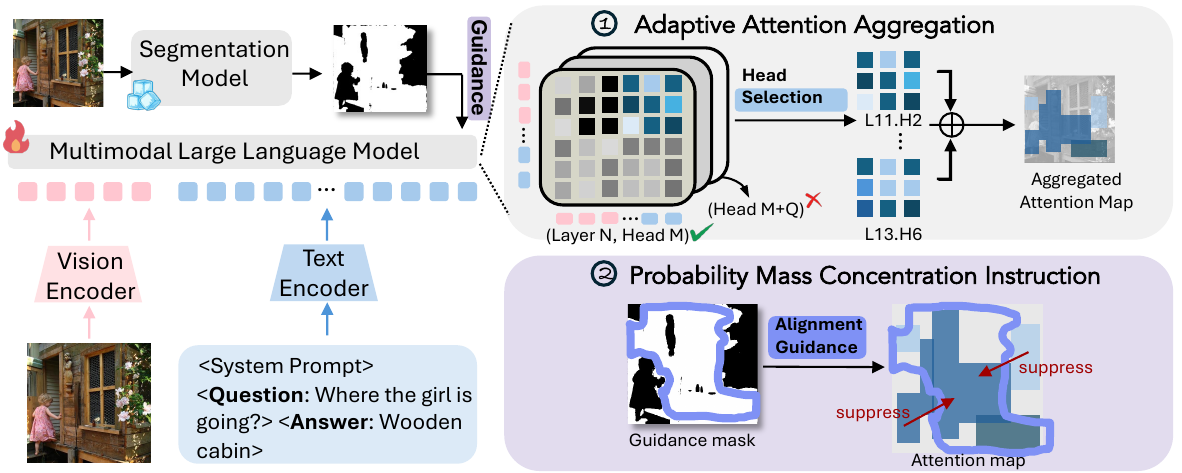} 
    \caption{Overview of the proposed sPMC framework. A text-conditioned segmentation model generates a binary guidance mask; Adaptive Attention Aggregation selects grounding-responsive heads, and probability-mass concentration directs their attention toward the masked regions. }
    \label{figure1}
\end{figure*}


\begin{itemize}
    \item We formulate region-level attention guidance as probability-mass concentration, providing flexible spatial supervision without exact patch-wise matching. 
    \item We introduce Adaptive Head Selection, which restricts probability-mass guidance to a small subset of attention heads with high visual engagement and show that selectively applying guidance through these heads is sufficient and more effective.
    \item We evaluate sPMC across six multimodal benchmark suites and multiple MLLMs, obtaining an average zero-shot gain of 3\% and gains of up to 11.3\% over the corresponding base models.
\end{itemize}

\section{Related Work}

\subsection{Multimodal Large Language Models}
MLLMs such as Qwen-VL~\cite{Bai2023Qwen-VL:Beyond}, InstructBLIP~\cite{Dai2023InstructBLIP:Tuning}, and LLaVA~\cite{Liu2024ImprovedTuning} achieve strong performance across multimodal tasks~\cite{Li2019VisualCaptioning,Deng2025ATask}. Many architectures connect a pretrained vision encoder, such as ViT~\cite{Dosovitskiy2020AnScale}, to a language model through a learned adapter. Decoder-only models such as Chameleon~\cite{Team2024Chameleon:Models} and Qwen3-VL~\cite{Bai2025Qwen3-VLReport} instead represent visual and textual inputs in a shared autoregressive sequence. In both designs, attention provides an important interface between queries and visual tokens, making it a natural target for grounding supervision.

\subsection{Enhancing Multimodal Reasoning}
Recent efforts to enhance multimodal reasoning in MLLMs largely focus on improving intermediate reasoning processes and grounding mechanisms to alleviate multimodal hallucinations. A prominent direction comprises Multimodal Chain-of-Thought (MCoT) methods~\cite{Zhang2024MultimodalModels,Gao2025Interleaved-ModalChain-of-Thought,Wu2024MindsModels}, which extend the success of CoT prompting in LLMs to multimodal settings by introducing explicit intermediate reasoning steps~\cite{Zheng2023DDCoT:Models, Chen2024VisualReasoning, Zhang2024MultimodalModels, Mitra2024CompositionalModels}. ICoT~\cite{Gao2025Interleaved-ModalChain-of-Thought}, for example, interleaves visual regions selected using attention scores, while MVoT~\cite{Li2025ImagineVisualization-of-Thought} pairs intermediate reasoning steps with generated visualizations. Such methods improve the reasoning process but may still inherit errors from the mechanism used to select visual evidence.

Another line of studies has identified cross-modal attention misallocation as a key factor contributing to the failure of visual reasoning tasks~\cite{Seil2025SeeModels, Woo2025DontModels, Chen2025Lavender:Tuning}. These works show that attention maps are often either overly diffuse or concentrated on irrelevant regions, leading to ineffective aggregation of visual evidence during reasoning. To address this issue, several approaches attempt to rectify attention by manipulating attention distributions. For instance, ~\cite{Seil2025SeeModels} applies diagnostic experiment to identify attention heads that consistently attend to irrelevant patches (attention sinks). They then propose an inference-time intervention that reallocates attention away from these sinks. \cite{Leng2024MitigatingDecoding} introduces contrastive decoding adjustments to refine attention allocation. Lavender~\cite{Chen2025Lavender:Tuning} enables MLLMs to learn towards diffusion's attention maps through patch-wise regression. While these techniques can partially improve visual perception by redistributing attention, their effectiveness is inherently limited, often yielding modest improvements and incurring additional inference overhead. Also, they fail to directly learn how attention is distributed across visual tokens.

\section{Methodology}
We propose a novel supervision framework to improve MLLMs’ general reasoning abilities by explicitly regularizing their spatial attention distribution. Built on the hypothesis that effective reasoning depends on concentrating attention over semantically relevant image regions, we first construct semantically enriched grounding masks using external segmentation priors. We then selectively identify attention heads that exhibit strong grounding behavior and aggregate multi-head, multi-layer attention into a unified spatial representation. Finally, we introduce a probability mass concentration objective that encourages the model to allocate more attention to relevant regions while preserving their pretrained generative capabilities. Together, these components provide a weakly supervised yet effective framework for aligning textual queries with visual evidence.
\subsection{Problem Formulation}

Let the visual encoder map an image $\mathcal{I}$ to $N_v$ visual tokens, let the language model autoregressively generate an answer $x=(x_1,\ldots,x_T)$ conditioned on $\mathcal{I}$ and a question $\mathcal{Q}$:
\begin{equation}
p_\theta(x\mid\mathcal{I},\mathcal{Q})
=
\prod_{t=1}^{T}
p_\theta(x_t\mid x_{<t},\mathcal{I},\mathcal{Q}).
\end{equation}
Our central hypothesis is that prediction errors on vision-centric tasks often arise when grounding-sensitive attention heads allocate insufficient attention to task-relevant visual evidence. We therefore formulate attention guidance as probability-mass concentration: selected heads are encouraged to allocate greater attention mass to relevant visual regions without being forced to match a fixed token-wise attention pattern.
\begin{algorithm}[t]
\small
\caption{Semantically Enriched Grounding Mask Generation}
\label{algorithm}
\begin{algorithmic}[1]
\Require Image $I$, caption or answer $T$, pretrained LLM $G$, text-conditioned segmentation model $S$, maximum foreground ratio $\rho_{\max}=0.6$
\Ensure Set of binary masks $\{\mathcal{M}_{j}\}$

\State Extract noun phrases with modifiers from $T$ to form proposals $\mathcal{P} = \{p_j\} \gets G(T)$

\ForAll{$p_j \in \mathcal{P}$}
    \State $\mathcal{R}_j = \{r_{i,j}\} \gets \text{S}(I, p_j)$ \Comment{candidate regions}
    
    \State $\mathcal{M}_j \gets \bigvee_{r_{i,j} \in \mathcal{R}_j} r_{i,j}$
    
    \State $\rho_j \gets \sum \mathcal{M}_j/{|I|}$ \Comment{foreground ratio}
    
    \If{$\rho_j < \rho_{max}$}
        \State Add $\mathcal{M}_j$ to output set
    \EndIf

\EndFor

\State \Return $\{\mathcal{P}= \{p_j\}, \mathcal{M}=\{\mathcal{M}_j\}\}$
\end{algorithmic}
\end{algorithm}
\subsection{Semantically Enriched Grounding Masks}\label{mask generation}
Dense attention annotations are rarely available at scale, so we derive weak spatial priors using SAM 3~\cite{Carion2026SAMConcepts}, which is designed to detect and segment visual concepts specified by short noun-phrase prompts.

Given an image, we prompt pre-trained LLMs to derive several semantically enriched text proposals by extracting noun phrases with language modifiers (e.g., color, size and locations) from captions or input context. The textual proposals specify not only object categories but also identifiable fine-grained properties that refer to a particular instance, such as \textit{``man wearing green shirt''} or \textit{``dog on the left''}. For caption-based samples, proposals are extracted from image-associated captions; for VQA samples, they are extracted from reference answers. We retain only concrete noun phrases that denote visually localizable entities. Because these source texts describe the corresponding image, this construction reduces the likelihood that a segmentation prompt refers to an absent concept. 

Conditioned on each proposal, SAM generates multiple candidate regions corresponding to different visual hypotheses. We then aggregate these segmentation masks into a single binary mask and discard masks with excessively large foreground ratio (>60\%) to avoid coarse or non-informative coverage. 
The resulting masks $\mathcal{M}_{i,j}\in \{0, 1\}$ capture semantically relevant regions corresponding to the noun phrase, which serve as region-level priors that guide text-to-image attention toward relevant areas, without enforcing overly strict supervision. Implementation details can be found in the Appendix and Algorithm~\ref{algorithm}. We treat the resulting masks as weak spatial priors rather than pixel-accurate ground truth. Merging candidate masks improves coverage, while the foreground-area filter removes overly broad masks. This favors recall because sPMC is more robust to moderate over-coverage than to missing relevant regions.
\subsection{Adaptive Attention Aggregation}
\subsubsection{Attention Aggregation.}

Let $\mathbf{Z}_{t}^{(h,l)}\in\mathbb{R}^{N_v}$ denote the pre-softmax text-to-image attention logits from answer token $t$ to the visual tokens, produced by head $h$ at layer $l$. Let $\mathcal{T}_{\mathrm{ans}}$ denote the token indices of the target answer span, and let $\mathcal{S}$ denote the set of head--layer pairs selected by the adaptive procedure described below. We aggregate the selected attention logits as
\begin{equation}
\tilde{\mathbf{Z}}
=
\frac{1}{|\mathcal{T}_{\mathrm{ans}}||\mathcal{S}|}
\sum_{t\in\mathcal{T}_{\mathrm{ans}}}
\sum_{(h,l)\in\mathcal{S}}
\mathbf{Z}_{t}^{(h,l)}.
\end{equation}
The corresponding spatial attention distribution is
\begin{equation}
\tilde{A}_i
=
\frac{\exp(\tilde{Z}_i)}
{\sum_{j=1}^{N_v}\exp(\tilde{Z}_j)},
\qquad i=1,\ldots,N_v.
\end{equation}
The distribution $\tilde{\mathbf{A}}\in\Delta^{N_v-1}$ is reshaped into the original $H_v\times W_v$ visual-token grid, where $H_vW_v=N_v$ (Figure~\ref{figure1}).

\begin{figure}[t]
    \centering \includegraphics[width=0.45\textwidth, height=130 pt]{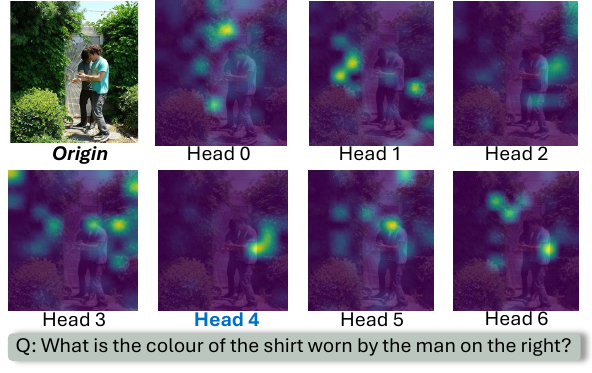} 
    \caption{Exemplar visualization of attention maps from layer 6 of Qwen3-VL-2B. While several heads exhibit irrelevant attention patterns, certain heads (Head 4) demonstrate strong alignment with the task-relevant region. }
    \label{figure2}
\end{figure}

\subsubsection{Adaptive Head Selection.} \label{adaptive selection rule}
Our exploratory analysis (Figure~\ref{figure2}) indicates that only a subset of attention heads produces spatially coherent and semantically grounded patterns, while others remain diffuse or specialized in text-only reasoning~\cite{George2025DifferentiationCoefficient, Seil2025SeeModels}. Uniformly regularizing all heads can disrupt non-grounding functions and dilute optimization signals. We therefore selectively regularize a target subset $\mathcal{H}_s \subseteq \mathcal{H}$ that exhibits strong visual engagement and spatial alignment. A head $h$ is included in $\mathcal{H}_s$ if it satisfies two criteria. The following head-level statistics are computed for each calibration example and then averaged over $\mathcal{D}_{\mathrm{cal}}$.


\paragraph{Criterion 1: Visual Engagement.}
Let \(Z_{t,i}^{(h,l)}\) denote the raw pre-softmax attention
logit from answer token \(t\) to visual token \(i\), produced
by head \(h\) at layer \(l\). We define the visual engagement
score as the mean logit over all answer--visual token pairs:
\begin{equation}
e^{(h,l)}
=
\frac{1}{|\mathcal{T}_{\mathrm{ans}}||\mathcal{V}|}
\sum_{t \in \mathcal{T}_{\mathrm{ans}}}
\sum_{i \in \mathcal{V}}
Z_{t,i}^{(h,l)} 
\label{eq:visual_engagement}
\end{equation}
A head--layer pair satisfies the visual-engagement criterion
if \(e^{(h,l)} \geq \tau_m\).



\paragraph{Criterion 2: Spatial Alignment.}
Spatial alignment measures the negative log attention mass assigned to the target region:
\begin{equation}
g^{(h,l)}
=
\frac{1}{|\mathcal{T}_{\mathrm{ans}}|}
\sum_{t\in\mathcal{T}_{\mathrm{ans}}}
\left(
\operatorname{LSE}_{i\in\mathcal{V}} Z_{t,i}^{(h,l)}
-
\operatorname{LSE}_{i\in\mathcal{R}} Z_{t,i}^{(h,l)}
\right)
\end{equation}
where $\operatorname{LSE}_{i\in\mathcal{A}}(u_i)
=\log\sum_{i\in\mathcal{A}}\exp(u_i)$. Lower values indicate stronger spatial alignment.

The selected set of head--layer pairs is
\begin{equation}
\mathcal{S}
=
\left\{
(h,l):
e^{(h,l)}\geq\tau_m
\ \text{and}\
g^{(h,l)}\leq\tau_s
\right\}
\end{equation}
The thresholds $\tau_m$ and $\tau_s$ are determined from the calibration statistics using the elbow method~\cite{Sahoo1988ATechniques} (detailed in Appendix).
\begin{table*}[th]
\small
\centering
\setlength{\tabcolsep}{4pt}
{
\begin{tabular}{l l l|c c c c c c c}
\toprule
Model & Base & Data 
& POPE
& HalluBench
& MMVP
& VisOnly$_S$
& VisOnly$_R$
& $V^*$ & MME \\

\midrule
\midrule
Qwen3-VL-2B & Qwen3-2B & N.A. 
& 89.0 & 44.1 & 51.3 &38.0&44.0& 67.5 & 2025.8\\
+ AutoR. LoRA-FT & Qwen3-2B & 0.3M 
& 84.5 &35.3 & 51.3 & 31.7&\textbf{46.2}&63.9&1525.0\\
\cellcolor{mycolor}+ sPMC LoRA-FT & \cellcolor{mycolor}Qwen3-2B & \cellcolor{mycolor}0.3M 
& \cellcolor{mycolor}\cellcolor{mycolor}\textbf{\underline{89.9}} & \cellcolor{mycolor}\textbf{49.1} & \cellcolor{mycolor}\textbf{52.7} & \cellcolor{mycolor}\textbf{38.3}&\cellcolor{mycolor}45.8&\cellcolor{mycolor}\textbf{68.1}&\cellcolor{mycolor}\textbf{2118.8} \\
\multicolumn{2}{l}{\textbf{\textit{\% improvement w.r.t. base model }}}&  & 1.0$\uparrow$
& 11.3$\uparrow$ & 2.7$\uparrow$ & 0.8$\uparrow$& 0.9$\uparrow$&1.8$\uparrow$ &4.6$\uparrow$ \\
\midrule

Qwen3-VL-8B & Qwen3-8B & N.A. 
& 88.0 & 58.5 & 65.3 & 42.2&54.6&73.8&\textbf{2416.2} \\

+ AutoR. LoRA-FT & Qwen3-8B & 0.3M 
& 86.5 & 56.6 & 64.0 & 40.3 &50.4&71.2&2317.5 \\

\cellcolor{mycolor}+ sPMC LoRA-FT & \cellcolor{mycolor}Qwen3-8B&\cellcolor{mycolor}0.3M 
& \cellcolor{mycolor}\textbf{88.5} & \cellcolor{mycolor}\textbf{\underline{59.5}} & \cellcolor{mycolor}\textbf{\underline{69.3}} & \cellcolor{mycolor}\textbf{\underline{43.1}}&\cellcolor{mycolor}\textbf{\underline{55.8}} &\cellcolor{mycolor}\textbf{\textbf{75.4}}&\cellcolor{mycolor}2400.5\\
\multicolumn{2}{l}{\textbf{\textit{\% improvement w.r.t. base model }}}& &0.6$\uparrow$ & 1.7$\uparrow$
& 6.1$\uparrow$ & 2.1$\uparrow$ & 2.2$\uparrow$ & 2.2$\uparrow$ &0.6$\downarrow$\\
\midrule
Llama3.2-11B-V & Llama3-11B & N.A. 
& 88.1 & 40.3 & 42.0 & 25.0  &33.4 &30.2 &1820.0\\
+ AutoR. LoRA-FT & Llama3-11B & 0.3M 
& 85.5 & 35.4 & 39.8 & 19.8&28.9&24.6&1736.6\\
\cellcolor{mycolor}+ sPMC LoRA-FT & \cellcolor{mycolor}Llama3-11B & \cellcolor{mycolor}0.3M 
& \cellcolor{mycolor}\cellcolor{mycolor}\textbf{88.5} & \cellcolor{mycolor}\textbf{40.5} & \cellcolor{mycolor}\textbf{44.0} & \cellcolor{mycolor}\textbf{25.0}&\cellcolor{mycolor}\textbf{37.0}&\cellcolor{mycolor}\textbf{32.5}&\cellcolor{mycolor}\textbf{1841.0} \\
\multicolumn{2}{l}{\textbf{\textit{\% improvement w.r.t. base model }}}&  & 0.5$\uparrow$& 0.5$\uparrow$ & 4.8$\uparrow$ & 0.0 & 10.8$\uparrow$&7.6$\uparrow$ &1.2$\uparrow$ \\
\midrule
\midrule
\multicolumn{9}{c}{\textbf{Small SOTA Models with Massive FT Data ($\geq$5M) or Large SOTA Models ($\geq$20B)}} \\

\midrule
Qwen2-VL-72B & Qwen2-72B & $\sim$50M 
& 87.2 & 58.1 & -- & 41.4&44.4 &-- & 2482.7 \\

Molmo-7B-O & Qwen2-7B & $\sim$35M 
& 86.7 & 42.5 & -- & 34.3&31.0&--&1714.0 \\

GPT-4V & N.A. & N.A. 
& 81.8 & 43.9 & 38.7 & 39.0&48.8&54.9&2070.2 \\

Gemini1.5-Pro & N.A. & N.A. 
& 88.2 & 55.9 & \textbf{40.7} & \textbf{42.4}&\textbf{52.6}&\textbf{71.7}&2110.6 \\
Llama-3.2-90B & Llama-3.1-70B & N.A.
& 86.3 & 44.1 & -- & 33.7 &37.1 &-- &1741.0\\
InternVL3-38B & Qwen-2.5-32B & N.A.
& \textbf{89.2} & \textbf{58.4} & -- & 39.0 &42.1 & --&\underline{\textbf{2500.7}}\\
\bottomrule
\end{tabular}
}
\caption{Zero-shot performance across multimodal benchmarks. The best within each group is shown in bold, and the overall best is \underline{underlined}. Fair evaluations are performed and supported by VLMEvalKit~\cite{Duan2024VLMEvalKit:Models}.}
\label{main_results}
\end{table*}

\subsection{Selective Probability Mass Concentration}

Let $\mathcal{R}$ denote the visual-token indices covered by the target mask. Given the aggregated logits $\tilde{\mathbf{Z}}$, the sPMC loss minimizes the negative log attention mass assigned to $\mathcal{R}$:
\begin{equation}
\mathcal{L}_{\mathrm{sPMC}}
=
-\log
\frac{\sum_{i\in\mathcal{R}}\exp(\tilde{Z}_i)}
{\sum_{i=1}^{N_v}\exp(\tilde{Z}_i)}
\end{equation}
Equivalently,
\begin{equation}
\mathcal{L}_{\mathrm{sPMC}}
=
\operatorname{LSE}_{i=1}^{N_v}(\tilde{Z}_i)
-
\operatorname{LSE}_{i\in\mathcal{R}}(\tilde{Z}_i)
\end{equation}
The objective depends on the total probability mass within $\mathcal{R}$ rather than on point-wise agreement with every masked token. Consequently, moderately over-inclusive masks do not force attention toward each incorrectly included token; the model can satisfy the objective by concentrating on any informative subset within the mask. This makes sPMC more tolerant to false-positive mask regions than point-wise alignment objectives. Our mask-generation procedure mitigates this risk by deriving prompts from image-associated text and merging candidate masks to favor coverage.

This objective increases the collective probability mass assigned to the target region without enforcing a fixed distribution among in-mask tokens. The final objective is
\begin{equation}
\mathcal{L}_{\mathrm{total}}
=
\mathcal{L}_{\mathrm{LM}}
+
\lambda\mathcal{L}_{\mathrm{sPMC}},
\end{equation}
where $\lambda$ controls the strength of attention regularization.
\section{Experiment}
\subsection{Experiment Setup}
\subsubsection{Grounding Mask Dataset Generation.} Following the method described earlier, we use Qwen3-Instruct-4B~\cite{Yang2025Qwen3Report} to extract important semantically enriched prompts from image captions, which serve as textual queries for subsequent grounding.
we apply SAM3~\cite{Carion2026SAMConcepts} to generate binary segmentation masks across two datasets: Flickr30k~\cite{BryanA.2015Flickr30kModels} which contains 31,783 images on human everyday activities with 158,915 captions; RLAIF-V 83k~\cite{Yu2025RLAIF-V:Trustworthiness}, which provides 83,132 image
pairs covering multiple sources (MSCOCO~\cite{Lin2015MicrosoftContext}, ShareGPT-4V~\cite{Chen2024ShareGPT4Video:Captions},
MovieNet~\cite{Huang2020MovieNet:Understanding}, VQA variants). Totally, 332,480 image-question-mask tuples are generated with our method. We evaluate the quality of these automatically generated masks in the Appendix.
\subsubsection{Multimodal benchmarks}
Our method is evaluated across six benchmark suites, reported as seven evaluation sets, grouped into two categories: general vision-language understanding and hallucination evaluation.  1) \textit{General vision-language understanding task} assesses comprehensive multimodal capabilities such as OCR and reasoning. We include four benchmarks under this category, including MMVP~\cite{Tong2024EyesLLMs}, VisOnly (Synthetic and Real splits)~\cite{Kamoi2025VisOnlyQA:Information}, V*~\cite{Wu2023V:LLMs} and MME~\cite{Fu2025MME:Models}.
 2) \textit{hallucination evaluation} includes POPE~\cite{Li2023EvaluatingModels} and HallusionBench~\cite{Guan2024HallusionBench:Models}, two specialized benchmarks designed to assess whether a model’s responses remain faithful to the input image, thereby evaluating its reliability.

\subsubsection{Baseline models}
As our method improves the model’s allocation of attention and can be simply applied to various MLLMs, we integrate it with the state-of-the-art Qwen3-VL family (Qwen3-VL-2B-Instruct and Qwen3-VL-8B-Instruct)~\cite{Bai2025Qwen3-VLReport} and Llama3.2-11B-Vision~\cite{Grattafiori2024TheModels} to verify the generalizability of the method.
\subsubsection{Implementation details} Experiments
are conducted on a single NVIDIA RTX Pro 6000 GPU. For fine-tuning, we employ LoRA~\citep{Hu2021LoRA:Models} to update only a subset of parameters. Training is performed under gradient accumulation over 4 or 8 steps (depending on the size of the base model) with an effective batch size of 32 using the AdamW optimizer~\citep{Loshchilov2017DecoupledRegularization}, with the learning rate set to $1e-4$. A cosine learning rate scheduler is applied, beginning with a warm-up phase of 100 steps with one hard reset in the middle of training. 
\begin{figure*}[th]
    \centering \includegraphics[width=.95\textwidth]{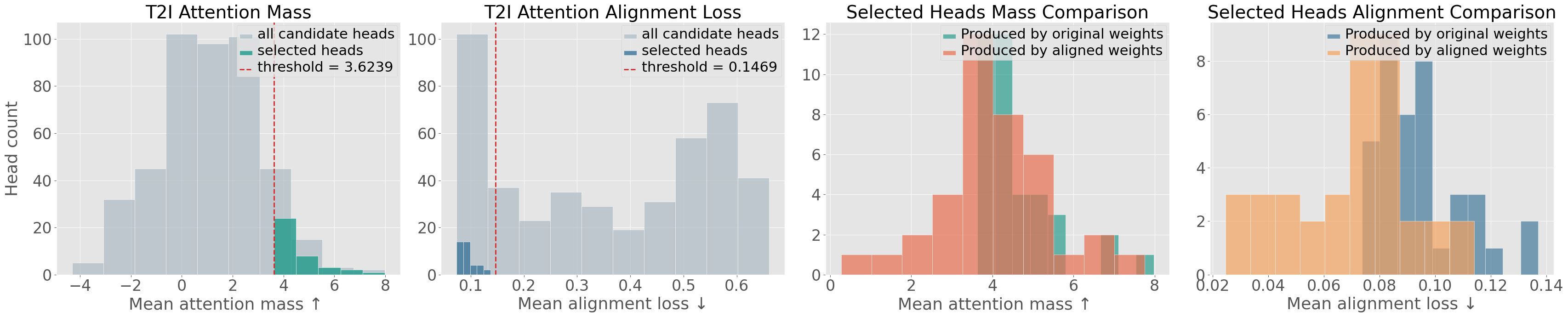} 
    \caption{
    Head-level diagnostics for Qwen3-VL-2B-Instruct, averaged over 200 randomly sampled images. The left panels show the calibration statistics and thresholds; the right panels compare selected heads before and after sPMC fine-tuning. Fine-tuning slightly reduces the mean visual-attention mass and shifts the alignment loss downward (lower is better).}
    \label{dist_shift}
\end{figure*}
\subsection{Main Results}
Table~\ref{main_results} presents a comprehensive comparison of zero-shot performance across multiple multimodal benchmarks under varying model scales and training regimes. Results demonstrate that the proposed method brings consistent performance improvements and reduced hallucination.

The results show that the proposed sPMC drives consistent performance gains on various baseline models across multiple benchmarks (on average 2.1\%-3.7\% depending on the base model). 
In particular, our proposed sPMC elevates Qwen3‑VL‑8B over existing state‑of‑the‑art performance, surpassing larger models such as GPT‑4V and Gemini 1.5 Pro on HallusionBench (59.5) and MMVP (69.3), despite being 7× smaller. This demonstrates that our method enables lightweight models to outperform their nominal capacity.
sPMC shows a specialized strength in reducing model hallucination. On two specialized hallucination challenges POPE and HallusionBench, our method yields average relative gains of 0.7\% and 4.5\% performance gains respectively to various base models, which verifies our theoretical hypothesis that better attention distribution can help alleviate MLLMs' hallucination.

\textbf{Effective selection of visual grounding attention heads.} Table~\ref{tab:head_selection_summary} and the two leftmost plots in Figure~\ref{dist_shift} reveal a consistent pattern in attention‑head selection across both decoder‑only and cross‑attention architectures. In all cases, only a small subset of heads is repeatedly selected, with selection rates ranging from 3.03\% to 15.4\%, indicating that visual grounding is highly sparse and concentrated in a limited number of heads. This sparsity is especially pronounced in larger models such as Qwen3‑VL‑8B, where only 3\% of heads are identified as relevant. Additionally, the decoder-only architecture rely only on self-attention mechanism, which entangles perception and reasoning within unified attention heads. This structural entanglement makes text-to-image attention guidance even more challenging for these more modern decoder-only models~\cite{Chen2025Lavender:Tuning}. Subsequently, it highlights the inherent significance of identifying grounding-relevant heads, as the results support restricting the alignment loss to heads most strongly associated with visual grounding. These observations motivate our design choice: instead of supervising all attention heads indiscriminately, focusing on a small subset of grounding-relevant heads enables more efficient and precise supervision. This insight underpins the proposed Adaptive Head Selection strategy.

\textbf{More grounded text-to-image attention.} The two rightmost plots in Figure~\ref{dist_shift}, which compare cross‑modal attention before and after applying sPMC, provide further evidence of the method’s effectiveness. After application of sPMC, the global distribution of text-to-image attention remains largely stable, with only a slight reduction in total attention mass, indicating that the model does not simply suppress attention. In contrast, the alignment loss decreases substantially, showing that the model reallocates attention toward task‑critical visual evidence. Together, these results demonstrate that sPMC sharpens text‑to‑image attention by preserving overall attention behavior while redirecting focus away from irrelevant regions and toward the correct visual cues.
\subsection{Ablation study}
\subsubsection{Adaptive head selection}
\begin{table}[t]
\small
\centering
\begin{tabular}{lccc}
\toprule
& \makecell{\textbf{Qwen3-}\\\textbf{VL-2B}}
& \makecell{\textbf{Qwen3-}\\\textbf{VL-8B}}
& \makecell{\textbf{Llama3.2-}\\\textbf{11B-V}} \\
\midrule
\textbf{\# Layers} 
& 28 & 36 & 8 \\
\midrule
\textbf{Total Heads} 
& 448 & 1152 & 256 \\
\midrule
\textbf{Selected Heads} 
& 69 & 35 & 37 \\
\midrule
\textbf{Selection Rate (\%)} 
& 15.40\% & 3.03\% & 14.45\% \\
\midrule
\end{tabular}
\caption{Summary of adaptive head selection, which demonstrates the sparsity of visual grounding heads. Detailed layer-wise head selection is provided in Appendix.}
\label{tab:head_selection_summary}
\end{table}
\begin{figure*}[t]
    \centering \includegraphics[width=.95\textwidth, height=120 pt]{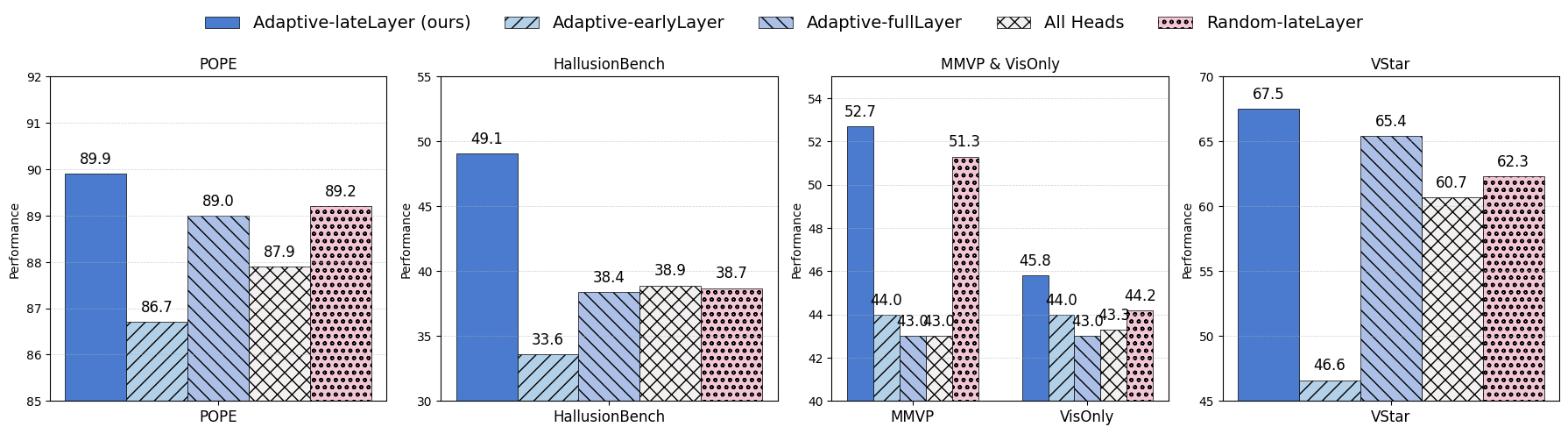} 
    \caption{Ablation study: zero-shot performance of head-selection variants for Qwen3-VL-2B-Instruct (28 layers and 448 heads). We partition layers into early (1--8), middle (9--16), and late (17--28) stages. Adaptive selection in late layers performs best on all five metrics.}
    \label{ablation2}
\end{figure*}

To evaluate Adaptive Head Selection, we compare \texttt{Adaptive-Late} with four variants: aggregation over all heads (\texttt{All Heads}), random selection from late layers (\texttt{Random-Late}), adaptive selection from early layers (\texttt{Adaptive-Early}), and adaptive selection across all layers (\texttt{Adaptive-Full}). All adaptive variants use the head selection criteria; \texttt{Adaptive-Late} applies them to layers 17--28.

\textbf{Late-layer guidance is more effective.} As shown in Figure~\ref{ablation2}, \texttt{Adaptive-Late} achieves the best performance across all five evaluation metrics. \texttt{All Heads} trails it by 6.24\% on average, indicating that indiscriminate regularization can dilute grounding-relevant signals. \texttt{Random-Late} is 3.9 points lower despite using the same layer range, showing that the gains depend on which heads are selected rather than merely on reducing their number. Layer placement is also important: \texttt{Adaptive-Early} trails \texttt{Adaptive-Late} by approximately 10 points, while \texttt{Adaptive-Full} improves upon \texttt{Adaptive-Early} by 4.9 points but remains inferior to late-layer selection. Together, these results support applying sPMC to adaptively selected grounding-responsive heads in later layers.

\begin{table}[th]
\small
\centering
\setlength{\tabcolsep}{3.5pt}
\begin{tabular}{l| c ccccc}
\toprule
\textbf{Method} &\textbf{Align}& \textbf{POPE} & \textbf{Hall.} & \textbf{MMVP} & \textbf{Vstar} & \textbf{Avg.} \\
\midrule
\midrule
\multicolumn{6}{c}{\textit{Qwen3-VL-2B}} \\
\midrule
No Guidance     & $\times$   & 89.0  & 44.1 & 51.3 &67.5 & 62.98\\
AutoR.     & $\times$ & 84.5 &35.3 & 51.3&63.9&58.75\\
CrossEntropy     & $\checkmark$ &  81.5&32.2  & 42.0 & 58.9 &53.65\\

\cellcolor{mycolor}\textbf{sPMC} & \cellcolor{mycolor}$\checkmark$&\cellcolor{mycolor}\textbf{89.9} & \cellcolor{mycolor}\textbf{49.1} & \cellcolor{mycolor}\textbf{52.7} & \cellcolor{mycolor}\textbf{68.1} & \cellcolor{mycolor}\textbf{64.95} \\

\midrule
\midrule
\multicolumn{6}{c}{\textit{Qwen3-VL-8B}} \\
\midrule
No Guidance     & $\times$   & 88.0  & 58.5 & 65.3 &73.8 & 71.40\\
AutoR.     & $\times$ & 86.5 &56.6 & 64.0&71.2&69.58\\
CrossEntropy     & $\checkmark$ &   81.8 & 45.6 & 57.3 &62.8&61.88\\

\cellcolor{mycolor}\textbf{sPMC} & \cellcolor{mycolor}$\checkmark$&\cellcolor{mycolor}\textbf{88.5} & \cellcolor{mycolor}\textbf{59.5} & \cellcolor{mycolor}\textbf{69.3} & \cellcolor{mycolor}\textbf{75.4} & \cellcolor{mycolor}\textbf{73.18} \\
\bottomrule
\end{tabular}
\caption{Ablation of the training objective. \texttt{AutoR.} denotes LoRA fine-tuning with only the next-token objective. Cross-Entropy directly matches attention to the normalized mask. The average is the unweighted mean of the four benchmark scores; $\times$ denotes no attention-alignment loss.}
\label{tab:sPMC_ablation}
\end{table}
\begin{figure*}[t]
    \centering \includegraphics[width=.95\textwidth, height=229 pt]{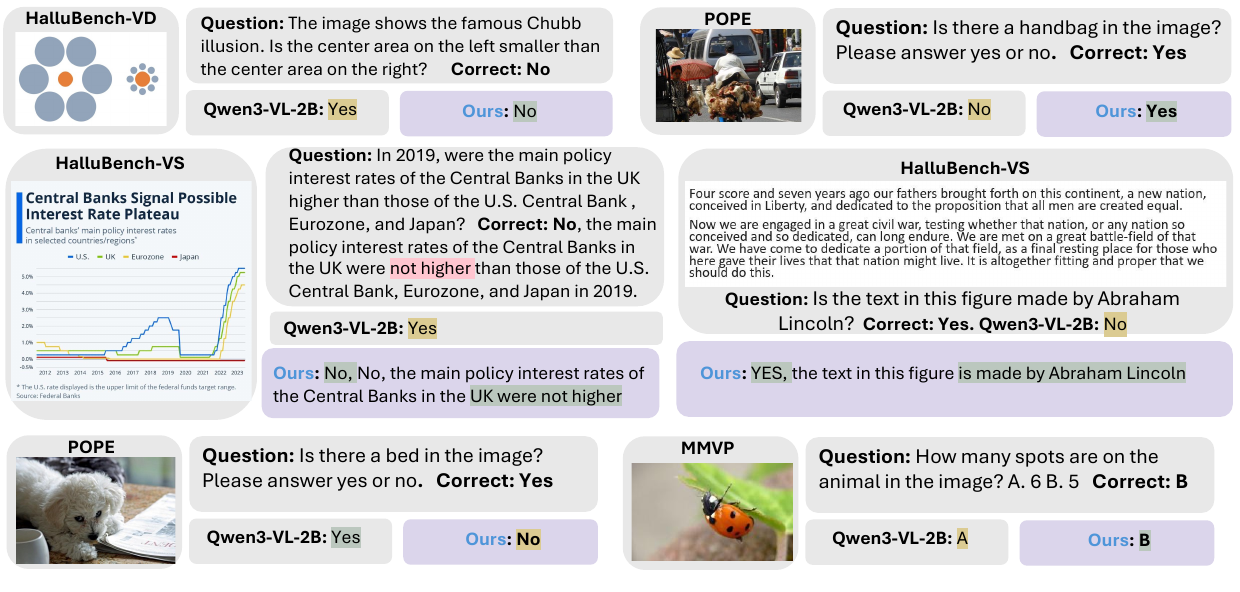} 
    \caption{Qualitative comparison between Qwen3-VL-2B and its sPMC-tuned counterpart. The examples cover diagram interpretation, OCR, commonsense reasoning, and fine-grained visual discrimination. Yellow marks incorrect answers and green marks correct answers. The bottom-left example is a failure case introduced by sPMC.}
    \label{examples}
\end{figure*}
\subsubsection{Training objective variants}
We compare sPMC with the pretrained model without fine-tuning (\texttt{No Guidance}), autoregressive LoRA fine-tuning (\texttt{AutoR.}), and patch-wise cross-entropy alignment with the normalized mask (\texttt{Cross-Entropy}).
The choice of alignment objective is critical. As shown in Table~\ref{tab:sPMC_ablation}, \texttt{AutoR.} and \texttt{Cross-Entropy} reduce average zero-shot performance by 3.03\% and 9.42\%, respectively, relative to the pretrained model. The larger degradation from \texttt{Cross-Entropy} indicates that exact patch-wise alignment imposes an overly restrictive attention target. In contrast, sPMC improves performance, supporting the use of flexible region-level probability-mass guidance over point-wise matching.
\subsection{Qualitative analysis}
We also present multiple qualitative examples in Figure~\ref{examples}. Across the presented examples, the performance gains are evident over various multimodal tasks, spanning from graphic/diagram interpretation (\textit{Chubb illusion} and \textit{Central bank} examples), OCR-intensive text understanding (the question about \textit{Abraham
Lincoln's quote}) to commonsense reasoning, and fine-grained visual-grounded discrimination. These improvements verify our method’s ability to steer attention toward task-relevant regions, thereby improving access to relevant visual evidence and reducing hallucination. 

An illustrative example in the top-right corner further highlights this advantage: the sPMC-tuned model demonstrates stronger alignment between salient visual evidence and generated responses, particularly in scenarios where critical cues are subtle or easily overlooked (e.g., \textit{the lady in the corner is carrying a handbag.}). By concentrating probability mass on relevant regions, the model is better at capturing fine-grained details that are essential for accurate reasoning.

However, the bottom-left example highlights a representative failure case: when task-relevant evidence is very ambiguous (e.g., \textit{a blurred pet bed in the background}), our model may under-attend to such low-saliency regions and fail to recover the correct answer. This suggests that while attention concentration improves robustness in most settings, it may also bias the model against weak signals, indicating a limitation in handling uncertain visual details. 

\section{Conclusion and Future Directions}
In this work, we study attention misallocation as a potential source of hallucination in MLLMs and reformulate it as a problem of suboptimal probability mass distribution over visual regions. To address this, we propose Selective Probability Mass Concentration (sPMC), an effective weak supervision framework that directly regularizes the global structure of cross-modal attention. By encouraging attention to concentrate on semantically relevant regions and selectively tuning grounding-relevant heads, sPMC enhances MLLMs' ``visibility'' while preserving the model’s inherent reasoning capabilities. Extensive experiments across multiple benchmarks demonstrate that our approach consistently enhances zero-shot performance over base models. A promising direction is to incorporate more diverse and challenging visual inputs, including synthetic data and compositional scenes. Such data could provide richer supervision and improve robustness under complex reasoning conditions.

\bibliography{main}

\lstset{%
	basicstyle={\footnotesize\ttfamily},
	numbers=left,numberstyle=\footnotesize,xleftmargin=2em,
	aboveskip=0pt,belowskip=0pt,%
	showstringspaces=false,tabsize=2,breaklines=true}
\floatstyle{ruled}
\newfloat{listing}{tb}{lst}{}
\floatname{listing}{Listing}

%
\pdfinfo{
/TemplateVersion (2027.1)
}

\setcounter{secnumdepth}{0} 

%


\title{Mind What Matters for Reasoning: Aligning Cross-Modal Attention via Selective Probability Mass Concentration}

\author{
    Written by AAAI Press Staff\textsuperscript{\rm 1}\thanks{With help from the AAAI Publications Committee.}\\
    AAAI Style Contributions by Peter Patel Schneider,
    Sunil Issar,\\
    J. Scott Penberthy,
    George Ferguson,
    Hans Guesgen,
    Francisco Cruz\equalcontrib\corresponding,
    Marc Pujol-Gonzalez\equalcontrib\corresponding
}
\affiliations{
    \textsuperscript{\rm 1}Association for the Advancement of Artificial Intelligence\\


    1101 Pennsylvania Ave, NW Suite 300\\
    Washington, DC 20004 USA\\
    proceedings-questions@aaai.org
%
}

\maketitle

\section{Preliminary Experiment}
We conduct a preliminary analysis to better understand the role of attention heads when taking on comprehensive multimodal tasks. Two key observations are: (1) within each model, only a small subset of attention heads consistently exhibits grounding behavior, characterized by relatively concentrated attention on semantically relevant visual regions. This suggests that visual evidence aggregation is not uniformly distributed across heads, but instead localized to a few specialized components.

(2) Second, this pattern is consistent across different model architectures and scales. Despite differences in design, we observe similar head-level behaviors, where a limited number of heads dominate grounding-related attention while others serve complementary roles such as contextual reasoning or diffusive exploring.

\begin{figure*}[tp]
    \centering \includegraphics[width=0.9\textwidth]{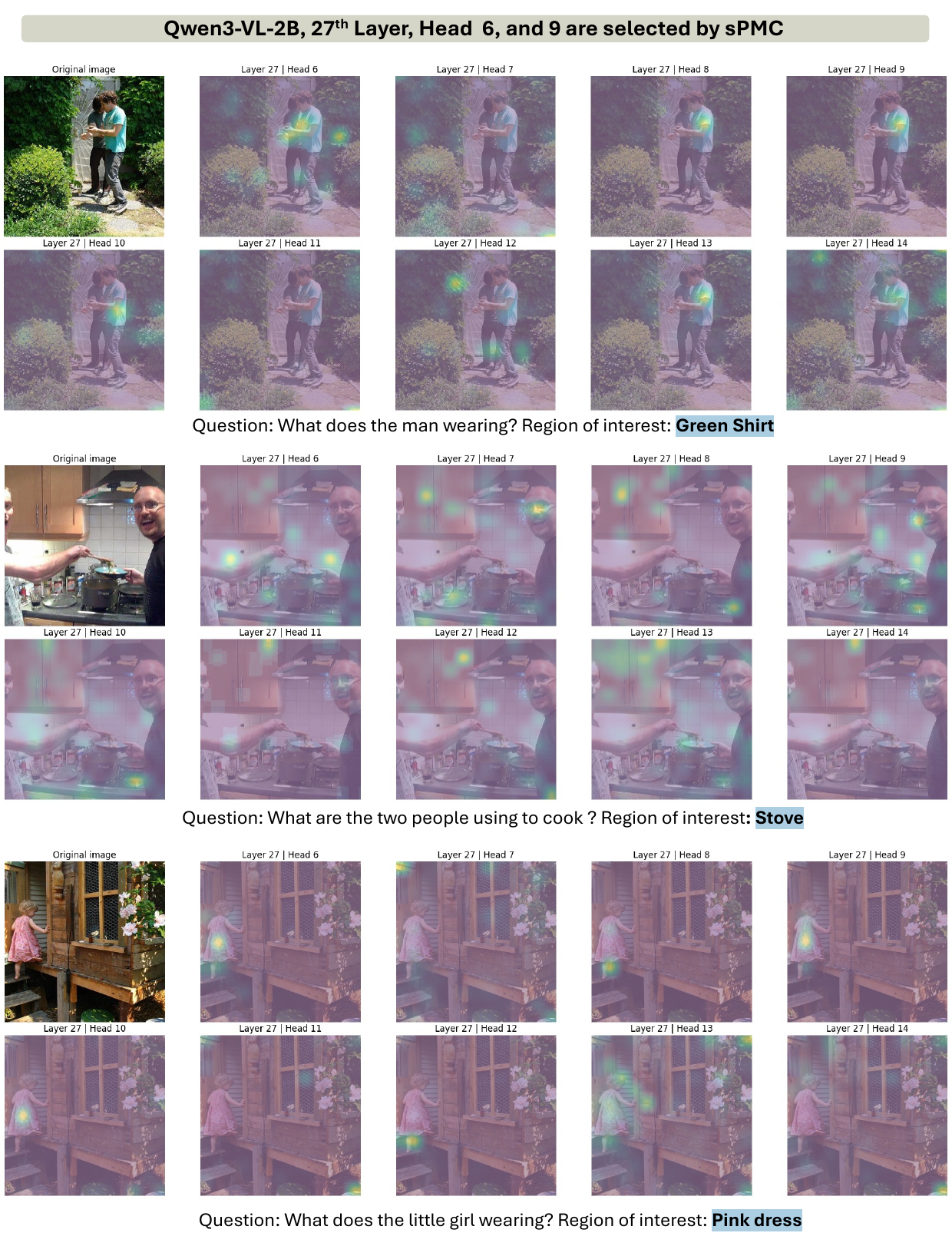} 
    \caption{Qualitative attention maps highlighting representative heads (e.g., Layer 27, head 6 and 9) that consistently capture relevant visual evidence across samples. Other head's attention (e.g., head 6, and 14) are scattered in the image or not focusing on the relevant region.  }
    \label{preliminary1}
\end{figure*}

\begin{figure*}[tp]
    \centering \includegraphics[width=0.9\textwidth]{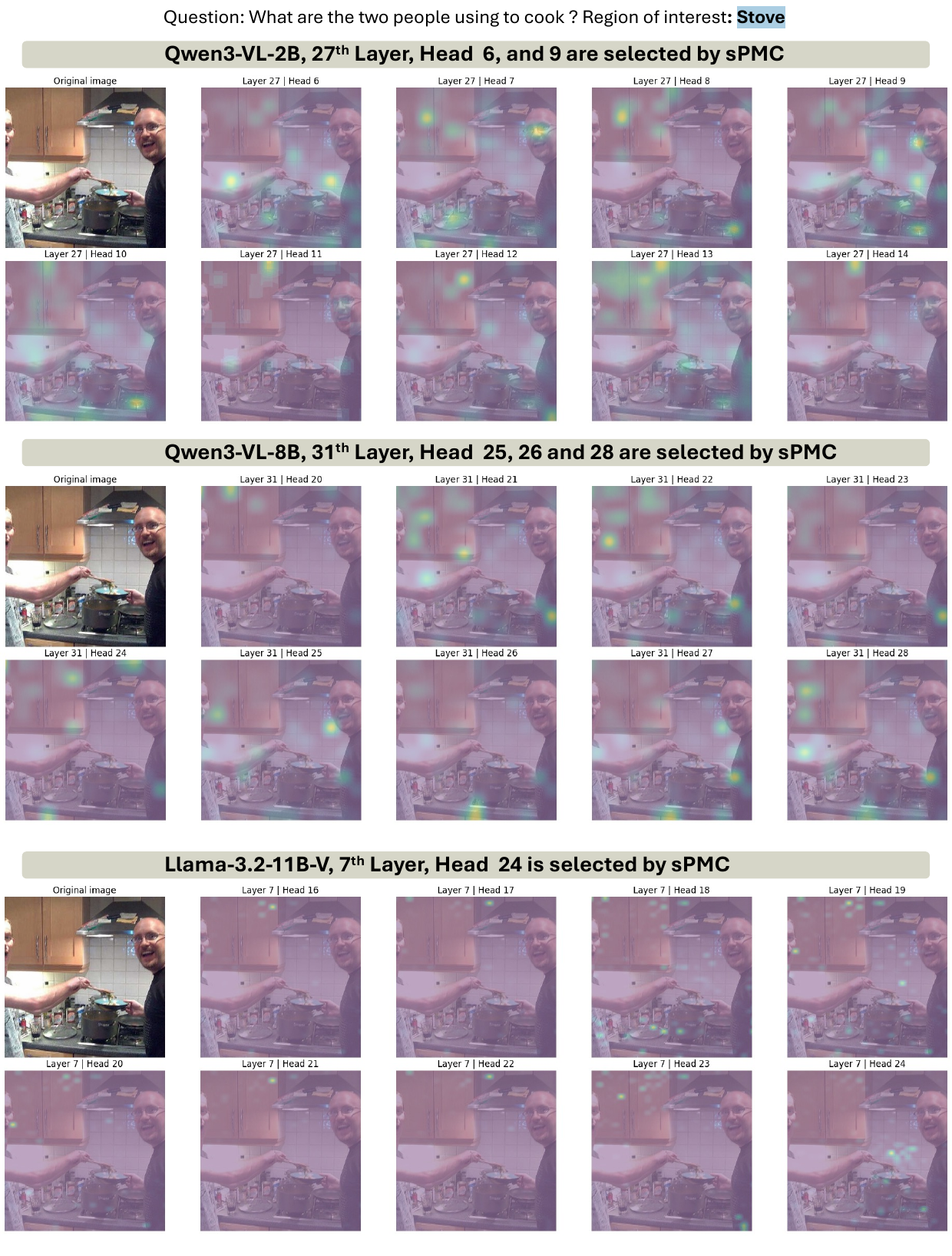} 
    \caption{Qualitative attention maps demonstrating visual evidence aggregation behavior exists across different model backbones: Qwen3-VL-2B, Qwen3-VL-8B, Llama-3.2-11B-V. For various backbones, the selected heads appeared to be more concentrated on relevant regions compared to other heads. }
    \label{preliminary2}
\end{figure*}

To support these observations, we provide two illustrative visualizations. Figure~\ref{preliminary1}: Qualitative attention maps that highlights representative heads that consistently capture visual grounding across samples. Figure~\ref{preliminary2}: qualitative attention maps that demonstrates visual evidence aggregation behavior existing across different model backbones. These results collectively support our motivation in method design that grounding is an emergent yet sparse property, motivating targeted supervision rather than uniform regulation across all heads.
\section{Details of Semantically Enriched Grounding Mask Generation}
This section describes the automated pipeline generating high-fidelity segmentation masks paired with noun phrases. This process transforms raw image-text pairs from the source dataset into grounded training data through a two-stage extraction and segmentation framework.

The \textit{initial stage} of our pipeline involves extracting concrete, segmentable entities from complex image descriptions. Given a caption from the dataset, we employ Qwen3-4B-Instruct~\cite{Yang2025Qwen3Report} as a semantic parser. This constraint ensures semantic alignment between the text and the image while filtering out abstract concepts (e.g., love, freedom) that are not suitable to localize. The output of this stage is a set of query-ready noun phrases $P = \{p_1, p_2, ..., p_n\}$, where each $p_i$ represents a localized entity with its original modifiers (e.g., colors, sizes, or spatial attributes). Modifiers turn isolated object nouns into fully grounded, semantically rich queries, which is critical for generating accurate masks that correspond to the intended object in context.The specific prompt we used is: 
\begin{tcolorbox}[
    colback=blue!3,
    colframe=blue!50!black,
    boxrule=0.6pt,
    arc=6pt,
    fonttitle=\bfseries,
    title=Prompt Design for Mask Generation,
    left=8pt,right=8pt,top=6pt,bottom=6pt
]
<SYSTEM PROMPT> Extract specific, individual physical objects or regions from the image description as noun or short noun phrases including key modifiers (e.g. color, size, location). Exclude categories and collections. The output must be literal substrings from the text. \\
\newline
<USER PROMPT> Refer to this image description: \{caption\}
\end{tcolorbox}

With the extracted semantic queries $P$, we perform zero-shot instance segmentation to generate binary spatial masks. We utilize the Segment Anything Model 3 (SAM3)~\cite{Carion2026SAMConcepts}, passing the image $I$ and the extracted phrase $p$ as a multimodal prompt pair. For each image-query pair, the model predicts binary mask of the same resolution as the image. 

To refine the spatial accuracy of our generated data, we configure the SAM3 inference parameters to balance semantic recall with spatial accuracy. We apply a detection threshold $\tau_{det} = 0.15$ to filter out low-confidence object proposals that lack sufficient alignment with the text prompt, and a mask binarization threshold $\tau_{mask} = 0.5$ to project the model's soft attention logits into discrete pixel-level assignments. This configuration allows the pipeline to capture challenging or small objects while establishing definitive decision boundaries for foreground pixels, ensuring that the resulting masks provide a sharp, unambiguous signal for downstream probability mass concentration tuning.

To maintain the integrity of the training data, we implement a semantic density filter to prune low-quality or overly-generalized masks. We define a foreground ratio $\rho$ as:$$\rho = \frac{1}{H \cdot W} \sum_{x,y} \mathcal{M}_{x,y}$$ Where $\mathcal{M}_{(x,y)}\in \{0,1\}$. $H$ and $W$ represent row and column sizes of the mask. We discard any mask where $\rho > 0.6$. This heuristic effectively filters out hallucinated segmentations or cases where the model fails to differentiate between a specific object and the global background. 

The resulting dataset provides a dense, semantically-grounded mapping that serves as the foundation for our visual reasoning training. Samples from the curated dataset are illustrated in Figure~\ref{sample_masks}. These examples demonstrate that the generated masks are semantically well-aligned with their corresponding queries and exhibit high spatial precision, with minimal annotation noise.


\section{Quality Evaluation of Segmentation Masks}
We conducted an AI-assisted semantic utility audit on 200 records sampled from the created dataset. For each record, a human and a blinded multimodal AI evaluator saw the original image, the target phrase, and a side-by-side visualization with the mask overlaid in translucent red; the source question and answer were withheld. The audit measures useful semantic localization for attention supervision rather than pixel-level segmentation accuracy. Imprecise boundaries, partial coverage that clearly localized the target, moderate over-coverage, and merged nearby regions were not automatically penalized. Each example was labeled Accurate, Usable, Unusable, or Unjudgeable. We release the audit artifact in the as Supplementary Media Archive. Each of the 200 records include image with segmentation mask, assessment on target visibility, target coverage, irrelevant coverage, confidence, and decision reason.

Overall, at least 195/200 examples (97.5\%) were determined Accurate or Usable. Excluding Unjudgeable cases, the usable-mask rate was 195/198 = 0.985 (98.5\%; Wilson 95\% CI [95.6\%, 99.5\%]). This quality is more than sufficient for the intended attention-supervision setting because the method is designed to be robust to noisy labels: masks act as weak spatial guidance, not pixel-level ground truth, and the small residual of noisy or ambiguous supervision can therefore be tolerated. 

\begin{figure*}[tp]
    \centering 
    \includegraphics[width=\textwidth]{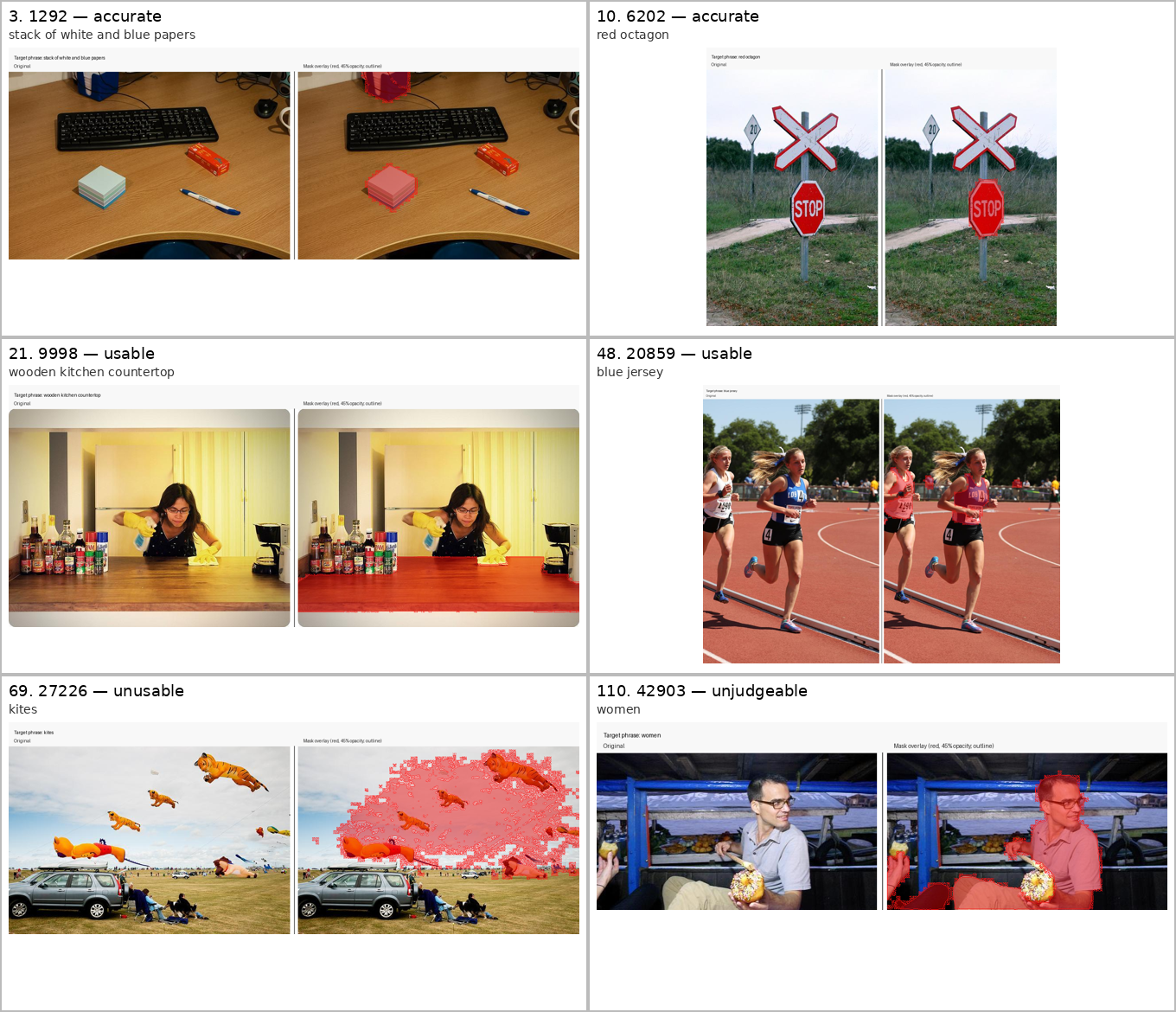} 
    \caption{Examples from our curated mask dataset and quality audit result. Red masks are the generated segmentation masks. }
    \label{sample_masks}
\end{figure*}

\begin{figure*}[tp]
    \centering 
    \includegraphics[width=\textwidth]{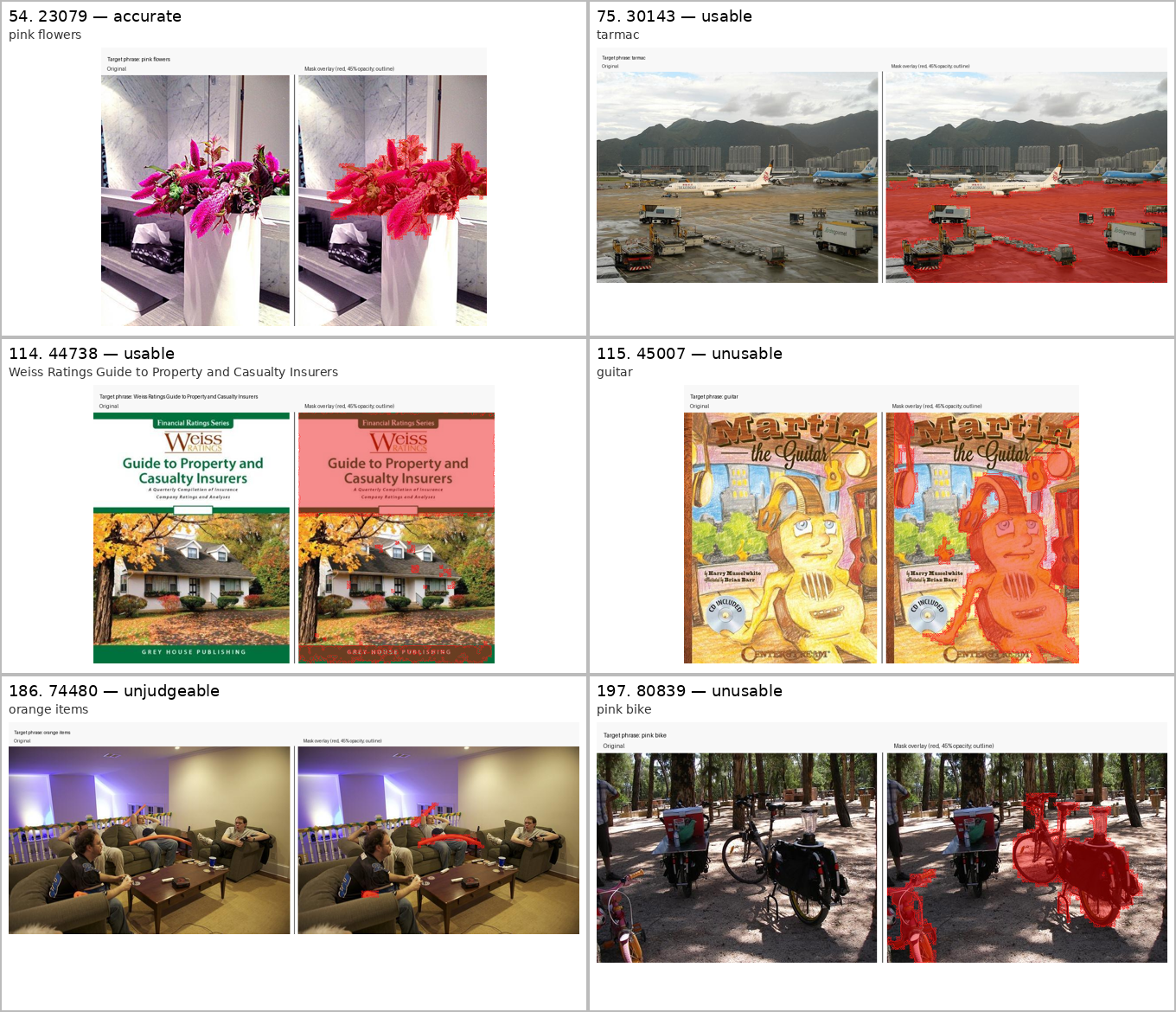} 
    \caption{Examples from our curated mask dataset and quality audit result (Continued).}
    \label{sample_masks_2}
\end{figure*}

Figure~\ref{sample_masks} and Figure~\ref{sample_masks_2} contain 12 examples from the evaluation: three clear Accurate masks, four imperfect but useful Usable masks, and \textbf{all} five Unusable/Unjudgeable cases identified by AI (human identified Unusable cases are a subset of the shown examples) for an extensive error analysis. It can be seen that the generated segmentation masks are generally quite accurate even for challenging cases. When focusing on the extremely rare error cases, the kite example is dominated by irrelevant sky; the ``women''example masks a man and foreground regions; the guitar example highlights the illustrated figure rather than the intended object; the pink-bike example focuses on other bicycles; and the broad ``orange items'' phrase is too ambiguous for reliable judgment. These cases show the main residual risks are wrong-target or over-global masks and ambiguous phrases. We do observe that the AI evaluator is slightly stricter than the human evaluator, especially around the criteria for Accurate and Usable. However, it does not change our conclusion that the quality of the generated segmentation mask set is sufficient for sPMC training.

\begin{table}[h]
\centering
\caption{Segmentation Mask Quality Audit Summary Statistics}
\label{tab:mask-quality-audit}
\begin{tabular}{lrrrr}
\toprule
Label & Human \% & Human \# & AI \% & AI \# \\
\midrule
Accurate & 77.0\% & 154 & 61.5\% & 123 \\
Usable & 22.0\% & 44 & 36.0\% & 72 \\
Unusable & 1.0\% & 2 & 1.5\% & 3 \\
Unjudgeable & 0\% & 0 & 1.0\% & 2 \\
\bottomrule
\end{tabular}
\end{table}


\section{Theoretical Justification of Adaptive Head Selection Criteria}
\subsection{Theoretical justification}
Based on our main formulation, we model each attention head can induce a conditional probability distribution over visual tokens:
\begin{align*}
    p_h(v|t)=\frac{\exp(a_{h,t}(v))}{\sum_{v'}\exp(a_{h,t}(v')}
\end{align*}

where $a_{h,t}(v)$ denotes the attention logit on visual token $v$ at decoding step $t$. Let $\mathcal{V}$ denote all visual tokens and $v' \subset \mathcal{V}$ is the semantically relevant region defined by the supervision mask.

\textit{\textbf{Criterion 1: High visual engagement.} Heads are selected that allocate substantial attention to visual tokens when processing target textual spans.}
\begin{align}
    \tilde{\mathcal{A}}(T_v)_h = \frac{1}{|T_{ans}|}\sum_{t\in T_{ans}}\mathcal{A}_{h}\geq\tau_m
\end{align}
Given the above conditional probability formulation, the criterion can be reformed as:
\begin{align}
    \tilde{\mathcal{A}}(T_v)_h = \frac{1}{|T_{ans}|}\sum_{t\in T_{ans}}\sum_{v\in \mathcal{V}}p_h(v|t)\geq\tau_m
\end{align}
Since cross-modal attention competes with textual tokens, this criterion enforces:
\begin{align}
    \mathbb E_t\left[
    \sum_{v\in \mathcal{V}}p_h(v|t)\right]\geq\tau_m
\end{align}
Which lower-bounds the marginal probability mass assigned to visual tokens. From an information-theoretic perspective, this ensures that the mutual information between the head and visual inputs is non-trivial. Thus excluding heads whose attention collapses onto textual tokens (i.e., $p_h(v)\approx 0$), which do not contribute to visual evidence aggregation.

\textit{\textbf{Criterion 2: Spatial Consistency.} Heads are selected when their attention sufficiently aligns with the visual region outlined by the supervision mask.}
\begin{align}
\mathbb{E}\left(\log\left(
1+
\exp\left(
\log \sum_{i,j} \exp\left(a^{-}_{i,j}\right)
-
\log \sum_{j} \exp\left(a^{+}_{i,j}\right)
\right)
\right)\right)\leq\tau_s
\end{align}
Let:
\begin{align}
Z^+ =\sum_{j} \exp\left(a^{+}_{i,j}\right) , Z^- = \sum_{i,j} \exp\left(a^{-}_{i,j}\right)
\end{align}
Then the above formula can be reduced to:
\begin{align}
    \mathbb E_t\left[\log(1+\exp(\log Z^- -\log Z^+)) \right]\leq\tau_s
\end{align}
This is precisely the negative log-likelihood of attention mass within the relevant region, which is implying:
\begin{align}
    \mathbb E_t\left[
    -\log\sum_{v\in \mathcal{V}^+}p_h(v|t)\right]\leq\tau_s
\end{align}
Which is equivalent to:
\begin{align}
    \mathbb E_t\left[
    \sum_{v\in \mathcal{V}^+}p_h(v|t)\right]\geq\delta(\tau_s)
\end{align}
It is noteworthy that we do not directly minimize $\log \frac{Z^-}{Z^+}$ in the practice as it is numerically unstable and unbounded. Therefore, we adopt the softplus transformation in practice as shown in Equation 4, which enables the smooth upper bound and computational stability while preserving the monotonicity of the original design.

Combining both criteria defines a feasible set $\mathcal{H}_s\subseteq\mathcal{H}$ that satisfies \textit{non-trivial visual engagement} and \textit{attending to semantically relevant regions.}

\subsection{Adaptive thresholds}
Rather than employing a fixed heuristic threshold, we dynamically determine the point of diminishing returns by analyzing the distribution of head scores. For a sorted set of scores $V$, we compute the optimal threshold by treating the distribution as a discrete curve and finding the point of maximum curvature relative to the curve's secant line. 

Given the sorted values $S = \{s_0, s_1, \dots, s_n\}$ and a chord $L$ connecting the endpoints $(0, s_0)$ and $(n, s_n)$, the optimal head index $k^*$ is defined as:$$k^* = \arg\max_{k} \frac{|(x_n - x_0)(y_0 - y_k) - (x_0 - x_k)(y_n - y_0)|}{\sqrt{(x_n - x_0)^2 + (y_n - y_0)^2}}$$where $x$ represents the rank and $y$ represents the score value.
This robustly ensures that the selected heads are statistically significant outliers in the performance distribution, resulting in a sparse but highly effective set of grounding-active heads.
\section{Experiment Setup Details}
\subsection{More Implementation Details}
We use parameter-efficient LoRA tuning and keep the remaining backbone parameters frozen. For all backbones, the adapters use rank 16, LoRA alpha 16, dropout 0.1, and no bias parameters. For Qwen3-VL-2B, LoRA is applied to the query and value projections in the first 14 of 28 language layers; for Qwen3-VL-8B, it is applied to the query and key projections in the final 15 of 36 language layers. For Llama-3.2-11B-Vision, LoRA is applied to the query and key projections of the cross-attention modules across the language backbone (i.e., only blocks containing those modules are affected). These architecture-specific placements are implementation choices rather than components of our method. We train each model for one epoch in bfloat16 with AdamW, a learning rate of $10^{-4}$, weight decay 0.1, 100 warmup steps, and a two-cycle cosine schedule with hard restarts. The effective batch size is 32 (using gradient accumulation where needed), and the random seed is 42.

\subsection{Benchmarks}
Our method is evaluated across multiple VLM benchmarks, grouped into two main categories: general vision-language understanding and visual hallucination task.  1) \textit{General vision-language understanding task} assesses comprehensive multimodal capabilities such as OCR and reasoning. There three benchmarks under this category: 
\begin{itemize}
    \item MMVP~\cite{Tong2024EyesLLMs} is a compact benchmark that contains 300 images-question pairs. It is designed to evaluate vision-language models on ``CLIP-blind" visual features. Higher performance on MMVP indicates the MLLMs can genuinely utilize visual information rather than relying on language priors.
    \item VisOnly (Synthetic and Real splits)~\cite{Kamoi2025VisOnlyQA:Information} evaluates MLLMs on 12 various geometric perception tasks, and reveals that base MLLMs often cannot accurately perceive basic geometric information in images. The dataset contains synthetic dataset (700 questions) and real dataset (900 questions).
    \item V*~\cite{Wu2023V:LLMs} is a visual search and grounding benchmark designed to evaluate Multimodal Large Language Models (MLLMs) on processing high-resolution images ($2246 \times 1582$ average resolution) and locating small visual details ($\text{area} < 0.05\%$). The dataset consists of 191 multiple-choice evaluation samples split into two core sub-tasks: attribute recognition (115 samples) and spatial relationship reasoning (76 samples).
    \item MME~\cite{Fu2025MME:Models} is a large-scale evaluation suite with over 14,000 evaluation samples spanning perception and cognition tasks. It includes fine-grained subtasks such as object recognition, OCR, commonsense reasoning, and numerical understanding, providing a comprehensive assessment of multimodal capability and robustness.
\end{itemize}

 2) \textit{Visual hallucination task.} In addition to assessing general MLLM capabilities, we conduct evaluations on two specialized benchmarks, where our method demonstrates strong effectiveness in mitigating hallucination:
 \begin{itemize}
     \item POPE~\cite{Li2023EvaluatingModels} is a diagnostic benchmark for probing object hallucination in MLLMs via binary (yes/no) questions about object presence. It comprises 3,000 questions constructed under different sampling strategies (e.g., random, popular, adversarial) to systematically test whether models can correctly assert the existence of certain objects.
     \item HallusionBench~\cite{Guan2024HallusionBench:Models} comprises 346 images paired with 1129 questions, all meticulously crafted by human experts. It is specifically designed to evaluate multimodal hallucination across diverse visual reasoning scenarios. 
 \end{itemize}
\subsection{Baseline models}
\begin{itemize}
    \item \textbf{Base Models.}We select Qwen3-VL-2B-Instruct, Qwen3-VL-8B-Instruct~\cite{Bai2025Qwen3-VLReport}, and Llama3.2-11B-Vision~\cite{Grattafiori2024TheModels} as our primary backbones to ensure diversity in both model scale and architectural design. Specifically, Qwen3-VL models represent strong \textbf{decoder-only self-attention} architectures at different parameter scales (2B vs 8B), enabling us to analyze scaling effects on attention behavior. In contrast, Llama3.2-11B-Vision adopts a \textbf{cross-attention-based design}, providing a complementary paradigm for multimodal fusion. This combination allows us to comprehensively evaluate the generality of our method across distinct attention mechanisms and model capacities.
    \item \textbf{Other Baseline Models.} We also consider alternative open-source MLLMs, including Qwen2-VL-72B~\cite{Bai2023Qwen-VL:Beyond}, InternVL3-72B~\cite{Zhu2025InternVL3:Models}, Molmo-7B-O~\cite{DeitkeMolmoModels} and Llama-3.2-90B~\cite{Grattafiori2024TheModels}, and close-source MLLMs, including Gemini1.5-Pro~\cite{Team2024GeminiContext} and GPT-4V. These models serve as reference points due to their leading performance on multimodal benchmarks, but are not directly comparable as they rely on larger scales (parameter size $\geq$ 20B) and complex post-training strategies (e.g., mixed preference optimization). In contrast, our method is evaluated on smaller MLLMs (2B-8B) under controlled settings, where it achieves comparable or even superior performance. Moreover, as a lightweight and model-agnostic approach, it can be readily integrated into larger models to further enhance their performance.
\end{itemize}
\begin{table}[h]
\centering
\begin{tabular}{lc}
\toprule
\textbf{Stage / Base} & \textbf{Computation Time} \\
\midrule
\textbf{Mask Generation}\\
\quad Noun Phrase Extraction & 0.04 s/iteration \\
\quad Grounding Mask Generation & 0.18 s/iteration \\
\midrule
\textbf{Training}  \\
\quad Qwen3-VL-2B & 2 h 25 min \\
\quad Qwen3-VL-8B & 4 h 43 min \\
\quad Llama3.2-11B-V & 7 h 26 min \\
\bottomrule
\end{tabular}
\caption{Computation time for semantically enriched grounding masks generation and model's instruction tuning.}
\label{tab:computation_time}
\end{table}
\begin{table*}[h]
\centering
\begin{tabular}{lccl}
\toprule
\textbf{Model} & \textbf{\#L} & \textbf{Sel. Rate (\%)} & \textbf{Selected Heads (by layer)} \\
\midrule
Qwen3-VL-2B$^*$ & 28 & \begin{tabular}[c]{@{}l@{}} 15.40\%\\(69/448) \end{tabular} & 
\begin{tabular}[c]{@{}l@{}}
L17: [6, 7, 8, 9, 10, 14, 15, 16]; \\L18: [2, 3, 9, 11, 12, 13]; \\ 
L19: [3, 4, 6, 12]; \\L20: [5, 6, 10, 14];\\ L21: [4, 5, 7, 9, 12, 13, 14, 15, 16]; \\ 
L22: [2, 9, 10, 13, 14, 16]; \\L23: [8, 10, 13]; \\L24: [1, 2, 5, 7, 15, 16]; \\ 
L27: [3, 5, 6, 9, 15]
\end{tabular} \\
\midrule
Qwen3-VL-8B$^*$ & 36 & \begin{tabular}[c]{@{}l@{}} 3.03\%\\(35/1152) \end{tabular} & 
\begin{tabular}[c]{@{}l@{}}
L22: [11, 19, 23, 24, 27];\\ L23: [23, 24]; L25: [9, 30, 31]; \\ 
L26: [31]; L29: [1, 24]; \\L30: [1, 4, 8, 13, 15]; \\ 
L31: [9, 10, 12, 25, 26, 28];\\ L32: [5, 6, 7, 8, 14, 15, 16]; \\ 
L33: [1, 11, 12]; L34: [24]
\end{tabular} \\
\midrule
Llama3.2-11B-V & 8 & \begin{tabular}[c]{@{}l@{}} 14.45\%\\(37/256) \end{tabular} & 
\begin{tabular}[c]{@{}l@{}}
L3: [9, 11, 25, 26, 27]; \\L4: [6, 8, 9, 10, 12, 21, 24, 25]; \\ 
L5: [1, 2, 3, 4, 5, 6, 7, 8, 13,\\ 16, 17, 18, 19, 20, 25, 27, 32]; \\ 
L6: [6, 9, 10, 12, 21, 23]; L7: [24]
\end{tabular} \\
\bottomrule
\end{tabular}
\caption{Full head selection across base models. Selection rate (Sel. Rate) is computed as the ratio of selected heads to total heads of the model. $^*$ indicates that cross-modal attention is achieved via self attention only. \#L means the total number of attention layers for each model. Selected head indexes are summarized in lists.}
\label{tab:head_selection_full}
\end{table*}
\section{More Experiment Results}
\subsection{Computation Time}
In terms of computational efficiency in Table~\ref{tab:computation_time}, our method introduces no additional overhead compared to prior approaches that explicitly modify or reweight attention maps during inference~\cite{Seil2025SeeModels, Woo2025DontModels,Leng2024MitigatingDecoding}. This design ensures that performance gains are achieved without sacrificing scalability or efficiency (training time within a few hours on a single GPU), making the method directly compatible with existing architectures and deployment settings.
\subsection{Adaptive Head Selection Results}
We summarize the details of head selection results in Table~\ref{tab:head_selection_full}.

\newpage


\end{document}